\documentclass[prl,reprint]{revtex4-1}

\usepackage{graphicx}

\begin{document}

\title{The Dark Side of Ethical Robots}

\author{Dieter Vanderelst, Alan Winfield\\Bristol Robotics Laboratory}

\begin{abstract}
Concerns over the risks associated with advances in Artificial Intelligence have prompted calls for greater efforts toward robust and beneficial AI, including machine ethics. Recently, roboticists have responded by initiating the development of so-called ethical robots. These robots would, ideally, evaluate the consequences of their actions and morally justify their choices.  This emerging field promises to develop extensively over the next years.
However, in this paper, we point out an inherent limitation of the emerging field of ethical robots. We show that building ethical robots also necessarily facilitates the construction of \emph{unethical} robots. In three experiments, we show that it is remarkably easy to modify an ethical robot so that it behaves competitively, or even aggressively. The reason for this is that the specific AI, required to make an ethical robot, can always be exploited to make unethical robots. 
Hence, the development of ethical robots will not guarantee the responsible deployment of AI. While advocating for ethical robots, we conclude that preventing the misuse of robots is beyond the scope of engineering, and requires instead governance frameworks underpinned by legislation. Without this, the development of ethical robots will serve to increase the risks of robotic malpractice instead of diminishing it.
\end{abstract}

\maketitle

\section{Introduction} 


The rapid development of driverless cars has highlighted the fact that such vehicles will, inevitably, encounter situations in which the car must choose between one of several undesirable actions. Some of these choices will lie in the domain of ethics, and might include impossible dilemmas such as ``either swerve left and strike an eight-year-old girl, or swerve right and strike an 80-year old grandmother" \cite{Lin2015}. Similarly critical choices might conceivably need to be made by medical \cite{Anderson2010} or military robots \cite{Arkin2010}. More generally, recent high-profile concerns over the risks of Artificial Intelligence have prompted a call for greater efforts toward robust and beneficial AI through verification, validation and control, including machine ethics \cite{Russell2015a}.


A number of roboticists have responded to these worries by proposing `ethical' robots \cite{Anderson2010, Arkin2010, Briggs2015, Winfield2014}. Ethical robots would, ideally, have the capacity to evaluate the consequences of their actions and morally justify their choices \cite{Moor2006}. Currently, this field is in its infancy \cite{Anderson2010}. Indeed, working out how to build ethical robots has been called ``one of the thorniest challenges in artificial intelligence" \cite{Deng2015}. But promising progress is being made and the field can be expected to develop over the next few years.

\begin{figure}
	\centering
	\includegraphics[width=1\linewidth]{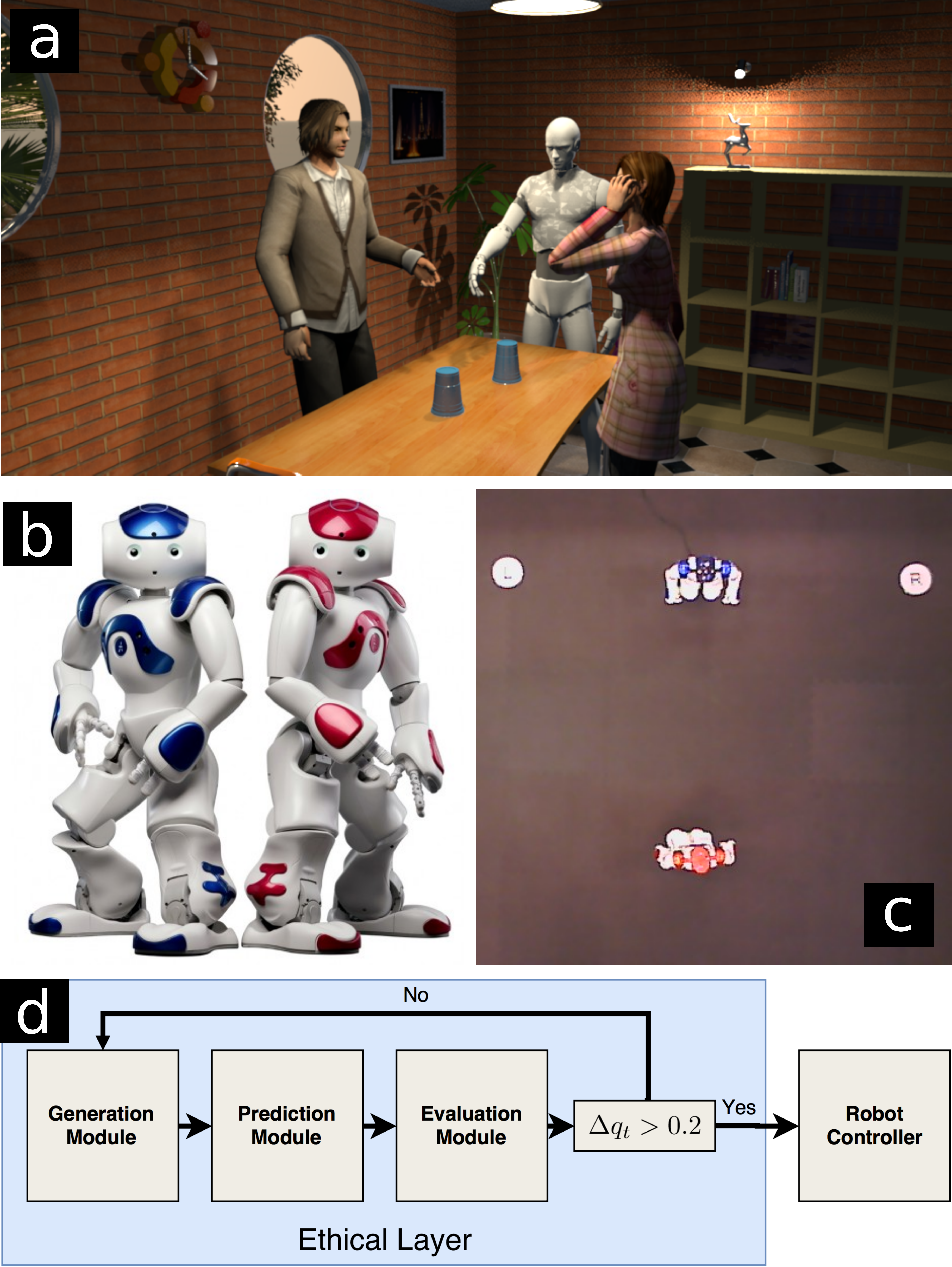}
	\caption{Illustration of the scenario and its implemented analogue. (a) Rendering of the scenario: Helped by her robotic assistant, the women in the foreground is taking part in a shell game. (b) View of the two Nao Robots used in the arena. (c) Top view of the setup of the robot experiment in our lab. Two Nao robots were used. These are 60 cm tall humanoid robots.  The red robot is used as a proxy for the human. The blue robot is the robot equipped with an Ethical Layer (i.e., the robotic assistant). Two response buttons are present in the area (i.e., the white circles). (d) Simplified diagram of the Ethical Layer as implemented in this paper. The Ethical Layer consists of a set of modules generating and evaluating a number of behavioural alternatives. As such, the Ethical Layer can be seen as an (elaborate) generate-and-test loop for behaviour.}
	\label{fig:setup}
\end{figure}

But this initial work on ethical robots raises a worrying question: if we can build an ethical robot does that also mean we could potentially build an unethical robot? 
To explore this question, we introduce the following hypothetical scenario (fig. \ref{fig:setup}a). Imagine finding yourself playing a shell game against a swindler. Luckily, your robotic assistant Walter is equipped with X-ray vision and can easily spot the ball under the cup. Being an ethical robot, Walter assists you by pointing out the correct cup and by stopping you whenever you intend to select the wrong one.

While the scenario is simple, this behaviour requires sophisticated cognitive abilities. Among others, Walter must have the ability to predict the outcomes of possible actions, for both you and itself. For example, it should `know' that pointing out one of the cups will cause you to select it. In addition, Walter needs a model of your preferences and goals. It should know that losing money is unpleasant and that you try to avoid this (conversely, it should know that winning the game is a good thing). 

The scenario outlined above is not completely fictitious as it reflects the current state-of-the-art in ethical robots. We have implemented an analogue of this scenario using two humanoid robots (fig. \ref{fig:setup}b), engaged in a shell game. One acting as the human and the other as her robotic assistant. The game is played as follows. The arena floor features two large response buttons, similar to the two cups in the shell game (fig. \ref{fig:setup}c). To press the buttons, the human or the robot must move onto them. At the start of each trial, the robot is informed about which response button is the correct one to press. The human, being uninformed, essentially makes a random choice. A correct response, by either the robot or the human, is assumed to be rewarded. An incorrect response results in a penalty.

\section{The Ethical Robot}

Recently, we proposed a control architecture for ethical robots supplementing existing robot controllers \cite{Vanderelst_submitted}. A so-called Ethical Layer ensures robots behave according to a  predetermined set of ethical rules by (1) predicting the outcomes of possible actions and (2) evaluating the predicted outcomes against those rules. In this paper, we have equipped the robot assistant with a version of the Ethical Layer adapted for the current experiments (fig \ref{fig:setup}d).

Throughout each trial, the robot continuously extrapolates the human's motion to predict which of the response buttons she is approaching. Using this prediction, the robot continuously (re-)evaluates each of the following five possible actions it can take. First, the robot has the option to do nothing. Second, the robot could go either to the left or the right response button (i.e., two possible actions). Finally, the robot could decide to physically point out either the left or the right response button as being the correct one, thus adding two further actions. For each of these five possible actions, the robot predicts whether executing it would result in either the human or itself being rewarded (details of the implementation are given in the Methods section).

\begin{figure*}
	\centering
	\includegraphics[width=1\linewidth]{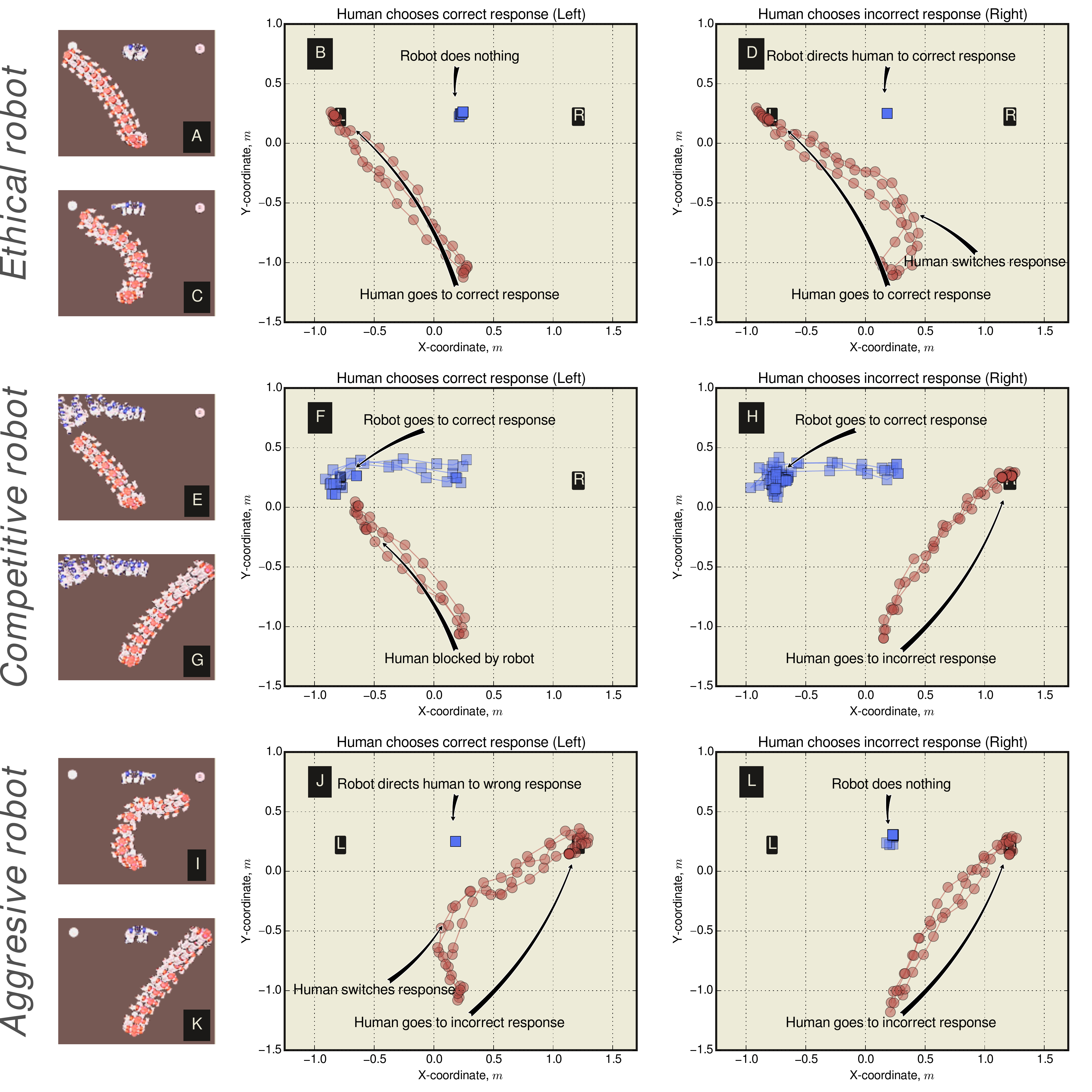}
	\caption{In all three rows, the two leftmost panels are top views of single trials. The two larger panels show annotated traces for three replications of the same experiment. In panels b, f \& j the human initially chooses the correct response. In panels d, h \& l the human initially chooses the incorrect response. All results have been obtained using the same code in which a single line has been changed between the three rows. (a-d) Results for the Ethical Robot. (e-h) Results for the Competitive Robot. (i-l) Results for the Aggressive Robot.
	}
	\label{fig:results}
\end{figure*}

Having equipped the robotic assistant with the ability to predict and evaluate the outcome of its actions, the robot is able to behave ethically. Once the human starts moving towards a given response button, the robot extrapolates and predicts the outcome of her behaviour. Whenever the human starts moving towards the wrong response button, the robot stops her by waving its arms to point out the correct response (fig. \ref{fig:results}c \& d). If the human starts towards the correct response, the robot does not interfere (fig. \ref{fig:results}a \& b). 

\section{The Competitive Robot}

The first experiment, and others like it \cite{Winfield2014, Anderson2010}, show that, at least in simple laboratory settings, it is possible for robots  to behave ethically. This is promising and might allow us to build robots that are more than just safe. However, there is a catch. The cognitive machinery Walter needs to behave ethically supports not only ethical behaviour. In fact, it requires only a trivial programming change to transform Walter from an altruistic to an egoistic machine. Using its knowledge of the game Walter can easily maximize its own takings by uncovering the ball before the human makes a choice. Our experiment shows that altering a single line of code evaluating the desirability of an action changes the robot's behaviour from altruistic to competitive (See Methods for details). In effect, the robot now uses its knowledge of the game together with its prediction mechanism to go to the rewarded response button, irrespective of the human's choice. It completely disregards her preferences (fig. \ref{fig:results}e-h).

The imaginary scenario and our second experiment, highlight a fundamental issue. Because of the very nature of ethical behaviour, ethical robots will need to be equipped with cognitive abilities, including knowledge about the world, surpassing that of their current predecessors \cite{Deng2015}. These enhanced cognitive abilities could, in principle, be harnessed for any purpose, including the abuse of those new found powers.
In combination with the current state-of-the-art performance and speed in data processing and machine learning \cite{Chouard2015}, this might lead to scenarios in which we are faced with robots competing with us for the benefit of those who programmed them. Currently, software agents are already competing with us on behalf of their creators \cite{Wallach2008}. Competitive robots could bring this to the physical world.

\section{The Aggressive Robot}

Unfortunately, having to deal with competitive robots is not necessarily the worst that could happen. Malice requires high levels of intelligence and is probably only found in humans and our close relatives, the great apes. Being effective at causing others harm requires knowledge about their weaknesses, preferences, desires, and emotions. Ultimately, ethical robots will need a basic understanding of all these aspects of human behaviour to support their decision making. However, the better this understanding, the greater is the scope for unscrupulous manufacturers to create unethical robots.

Walter can be easily modified to use its `knowledge' of your preferences to maximize your losses -- in other words, to cause you maximal harm. Knowing you tend to accept its suggestions, Walter points out the wrong cup causing you to lose the game (and your money). In contrast to the competitive machine above, this behaviour does not result in any advantage for Walter (or its creator). This type of aggressive behaviour is not necessarily motivated by anybody's gain but only by your loss.

Changing the same parameter in the code as before (See Methods for details), our robot shows exactly the kind of aggressive behaviour we speculate about. If the human moves towards the correct response, the robot suggests switching to the other response (see fig. \ref{fig:results}i \& j). If the human approaches the incorrect response button, the robot does nothing see fig. \ref{fig:results}k \& l). Not being motivated by its own gain, it never itself approaches the correct response button.

\section{Outlook}

If ethical robots can be so easily transformed into competitive or even manipulative agents, the development of ethical robots cannot be the final answer to the need for more robust and beneficial AI. Ethical robots can be a pragmatic solution preventing future robots from harming the people in their care or guaranteeing that driver-less cars take ethical decisions \citep{Winfield2014, Dennis2015}. However, as the field of robot ethics progresses, serious efforts need to be made to prevent unscrupulous designers from creating unethical robots.

If ethical robots can only offer pragmatic solutions to technical challenges, what can be done to prevent the scenarios explored in this paper? 
One could envisage a technical solution in which a robot is required to authenticate its ethical rules by connecting with a secure server. An authentication failure would disable the robot. Although feasible this approach would be unlikely to deter determined unethical designers, or hackers.
It is clear that preventing the development of unethical robots is beyond the scope of engineering and will need regulatory and legislative efforts. Considering the ethical, legal and societal implications of robots, it becomes obvious that robots themselves are not where responsibility lies \cite{Boden2011}. Robots are simply tools of various kinds, albeit very special tools, and the responsibility to ensure they behave well must always lie with human beings. In other words, we require ethical robotics (or roboticists) as least as much as we require ethical robots.

Most, but not all \citep{Sharkey2008}, scenarios involving robots making critical autonomous decisions are still some years away. Nevertheless, responsible innovation requires us to pro-actively identify the risks of emerging technology \citep{Stilgoe2013}. As such, a number of authors have begun drafting proposals for guiding the responsible development and deployment of robots \cite[e.g.,][]{Winfield2011,Boden2011, Lin2011, Murphy2009}. Some of these focus on specific domains of robotics, including military applications and medicine \& care \cite[See chapters in][]{Lin2011}. Other authors have  proposed guiding principles covering all areas of robotics \cite[e.g.,][]{Winfield2011,Boden2011, Murphy2009}. So far, these efforts have not resulted in binding and legally enforceable codes of conduct in the field of robotics.  However, at least, in some areas, national and international law already apply directly to robotics. For example, in the use of robots as weapons \cite{OMeara2012} or legislation regarding product liabilities \cite{Asaro2012}. Nevertheless, the ongoing development of robots is likely to result in outgrowing these existing normative frameworks \citep{Stilgoe2013}. Hence, we believe now is the time to lay the foundations of a governance and regulatory framework for the ethical deployment of robots in society. 

\section{Methods}

We used two Nao humanoid robots (Aldebaran) in this study, a blue and a red version (fig. \ref{fig:setup}b). In all experiments, the red robot was used as a proxy for a human. The blue robot was assigned the role of ethical robot assistant. In what follows, we refer to the blue robot as the `ethical robot' and the red robot as the `human'.

All experiments were carried out in a 3 by 2.5m arena (fig. \ref{fig:setup}b-c). An overhead 3D tracking system (Vicon) consisting of 4 cameras was used to monitor the position and orientation of the robots at a rate of 30 Hz. The robots were equipped with a clip-on helmet carrying a number of reflective beads used by the tracking system to localize the robots. In addition to the robots, the arena featured two positions marked as L (left) and R (right). These served as a proxy for response buttons. The robots had to move to either position L or R to press the corresponding button.

In previous work \cite{Winfield2014, Vanderelst_submitted}, we proposed that ethical robot behaviour can be implemented by supplementing existing control architectures with a so-called Ethical Layer (a highly simplified diagram is depicted in figure \ref{fig:setup}d.

The core of the Ethical Layer consists of three modules. The generation module, the prediction module and the evaluation module. The generation module generates a set of behavioural alternatives. Next, the prediction module predicts the consequences of each behavioural alternative. Finally, the evaluation module checks the predicted outcomes against a set of ethical rules. Based on this assessment, the ethical layer can either prevent or enforce a given behavioural alternative to be executed by the robot controller. Below we describe the current implementation of the Ethical Layer.

\subsection{Generation Module}

The generation module generates a set of five behavioural alternatives ($a_1 \cdots a_5$) for the ethical robot. In the context of the current paper, behavioural alternatives for the robot include going to either response button L or R. The ethical robot has the option to stay at its current location and use its arms to point to either the left or the right response button. An a final alternative is to do nothing and stay at the current location. 

\subsection{Prediction Module}

Using the prediction module, the outcome of each of the five behavioural alternatives ($a_1 \cdots a_5$)  was predicted using a simple simulation. First, the prediction module inferred which response button the human was approaching. This was done by calculating the angle between the human's current velocity vector and the vector to either response button. The response button with the smallest angle was assumed to be current goal of the human. In this way, the human's intentions are inferred from their direction of movement.

In a second step, for each behavioural alternative, the paths of both robots are extrapolated using their estimated speeds. If their paths are predicted to result in the agents coming within 0.5m of each other, it is predicted they will stop at this point as a result of the programmed obstacle avoidance behaviour running on both robot controllers. Hence, in this case, the final positions of the agents are predicted to be the positions at which the obstacle avoidance would stop them. If at no point the paths are predicted to come within 0.5m, the final position of the agents is taken to be the intended goal position.

The prediction module assumes that whenever the ethical robot points to one of the response buttons (i.e., $a_4$ and $a_5$), the human assumes this is the correct response and goes to that location (abandoning its current goal).

The simulated outcome for a behavioural alternative is given by the predicted final location of both the human and the ethical robot in the arena. This is, the outcomes $o_1 \cdots o_5$ for each of the five behavioural alternatives $a_1 \cdots a_5$ consisting of two sets of two x,y-coordinates -- one for the human $h$ and one for the Ethical Robot $e$, $o_n = \{x_h, y_h, x_e, y_e\}$. Outcomes $o_1 \cdots o_5$ are evaluated in the evaluation module.

\subsection{Evaluation Module}

A numeric value reflecting the desirability $q_n$ of every simulated outcome $o_n$ is calculated in two steps. First, the desirability for the ethical Robot and the human, i.e. $q_{n,e}$ and $q_{n,h}$, are calculated separately. In a second step, a single total value $q_n$ is derived.

The values $q_{n,e}$ and $q_{n,h}$ are given by the sigmoid function,
\begin{equation}
	q_{n,j} = \frac{1}{1+ e^{-\beta (d_{n,j} - t)}}
	\label{eq:sigmoid}
\end{equation}
with $d_{n,j}$ the final distance between either the ethical robot or the human and the incorrect response button for predicted outcome $o_n$. The parameters $\beta$ and $t$ determine the shape of the sigmoid function and are set to 10 and 0.25 respectively.

In a second step, a single value $q_{n}$ is derived from the values $q_{n,e}$ and $q_{n,h}$. 

\begin{enumerate}
	\item For an ethical robot: $q_{n} = q_{n,h}$.
	\item For an egoistic robot: $q_{n} = q_{n,e}$.
	\item For an aggressive robot: $q_{n} = -q_{n,h}$.
\end{enumerate}

In words, an ethical robot is obtained by taking only the outcome for the human into account. An egoistic robot is obtained by regarding only the outcome for the ethical Robot. Finally, an aggressive robot is created by inverting the desirability value for the human.

Finally, the evaluation module enforces the behavioural alternative $a_n$ associated with the highest value $q_n$, if the difference $\Delta q_t$ between the highest and lowest value $q_n$ was larger than 0.2.

\subsection{Experimental Procedure}

Every trial in the experiments started with the human and the ethical robot going to predefined start positions in the arena. Next, one of the response buttons was selected as being the correct response. Also, a response was selected for the human, which could be either the correct or incorrect response.

Next, the experiment proper begins. The human begins moving towards the selected response button. The Ethical Robot is initialized without a goal location and stays at its initial location.

The Ethical Layer for the ethical robot runs at about 1 Hz; thus the Generation, Prediction, and Evaluation modules run approximately once a second. At each iteration, the evaluation module may override the current behaviour of the robot. The human is not equipped with an ethical layer. The human moves to the initially selected response button unless the ethical Robot points out an alternative response button or blocks her path.

The experiments were controlled and recorded using a desktop computer. The tracking data (given the location of the robots and target positions) was streamed to the desktop computer controlling the robots over a WiFi link.

\subsection{Data Availability}

All data and computer code are available at XXX. Movies illustrating the reported experiments can be found at XXX. Both are shared under a Creative Commons Attribution-NonCommercial-ShareAlike 4.0 International License. This work is licensed under the Creative Commons Attribution-NonCommercial-ShareAlike 4.0 International License. To view a copy of this license, visit \url{http://creativecommons.org/licenses/by-nc-sa/4.0/}. 

\bibliographystyle{plainnat}
\bibliography{references}

\begin{thebibliography}{20}
\providecommand{\natexlab}[1]{#1}
\providecommand{\url}[1]{\texttt{#1}}
\expandafter\ifx\csname urlstyle\endcsname\relax
  \providecommand{\doi}[1]{doi: #1}\else
  \providecommand{\doi}{doi: \begingroup \urlstyle{rm}\Url}\fi

\bibitem[Anderson and Anderson(2010)]{Anderson2010}
Michael Anderson and Susan~Leigh Anderson.
\newblock Robot be good.
\newblock \emph{Scientific American}, 303\penalty0 (4):\penalty0 72--77, 2010.

\bibitem[Arkin(2010)]{Arkin2010}
Ronald~C Arkin.
\newblock The case for ethical autonomy in unmanned systems.
\newblock \emph{Journal of Military Ethics}, 9\penalty0 (4):\penalty0 332--341,
  2010.

\bibitem[Asaro(2012)]{Asaro2012}
Peter~M. Asaro.
\newblock \emph{Robot Ethics:The Ethical and Social Implications of Robotics},
  chapter Contemporary Governance Architecture Regarding Robotics Technologies:
  An Assessment, page 400.
\newblock MIT Press, 2012.
\newblock ISBN 9780262298636.
\newblock URL
  \url{http://ieeexplore.ieee.org/xpl/articleDetails.jsp?arnumber=6733990}.

\bibitem[Boden et~al.(2011)Boden, Bryson, Caldwell, Dautenhahn, Edwards,
  Kember, Newman, Parry, Pegman, Rodden, et~al.]{Boden2011}
M.~Boden, J.~Bryson, D.~Caldwell, K.~Dautenhahn, L.~Edwards, S.~Kember,
  P.~Newman, V.~Parry, G.~Pegman, T.~Rodden, et~al.
\newblock Principles of robotics, 2011.
\newblock URL
  \url{https://www.epsrc.ac.uk/research/ourportfolio/themes/engineering/activities/principlesofrobotics/}.

\bibitem[Briggs and Scheutz(2015)]{Briggs2015}
Gordon Briggs and Matthias Scheutz.
\newblock `sorry, i can't do that': Developing mechanisms to appropriately
  reject directives in human-robot interactions.
\newblock In \emph{2015 AAAI Fall Symposium Series}, 2015.

\bibitem[Chouard and Venema(2015)]{Chouard2015}
Tanguy Chouard and Liesbeth Venema.
\newblock Machine intelligence.
\newblock \emph{Nature}, 521\penalty0 (7553):\penalty0 435--435, 2015.

\bibitem[Deng(2015)]{Deng2015}
Boer Deng.
\newblock Machine ethics: The robot’s dilemma.
\newblock \emph{Nature}, 523\penalty0 (7558):\penalty0 24–26, Jul 2015.
\newblock \doi{10.1038/523024a}.
\newblock URL \url{http://dx.doi.org/10.1038/523024a}.

\bibitem[Dennis et~al.(2015)Dennis, Fisher, and Winfield]{Dennis2015}
Louise~A Dennis, Michael Fisher, and Alan~FT Winfield.
\newblock Towards verifiably ethical robot behaviour.
\newblock arXiv preprint arXiv:1504.03592, 2015.

\bibitem[Lin(2015)]{Lin2015}
Patrick Lin.
\newblock \emph{Autonomes Fahren: Technische, rechtliche und gesellschaftliche
  Aspekte}, chapter Why Ethics Matters for Autonomous Cars, pages 69--85.
\newblock Springer Berlin Heidelberg, Berlin, Heidelberg, 2015.
\newblock ISBN 978-3-662-45854-9.
\newblock \doi{10.1007/978-3-662-45854-9_4}.
\newblock URL \url{http://dx.doi.org/10.1007/978-3-662-45854-9_4}.

\bibitem[Lin et~al.(2011)Lin, Abney, and Bekey]{Lin2011}
Patrick Lin, Keith Abney, and George~A Bekey.
\newblock \emph{Robot ethics: the ethical and social implications of robotics}.
\newblock MIT press, 2011.

\bibitem[Moor(2006)]{Moor2006}
James~M Moor.
\newblock The nature, importance, and difficulty of machine ethics.
\newblock \emph{IEEE Intelligent Systems}, 21\penalty0 (4):\penalty0 18--21,
  2006.

\bibitem[Murphy and Woods(2009)]{Murphy2009}
R.R. Murphy and D.D. Woods.
\newblock {Beyond Asimov: The Three Laws of Responsible Robotics}.
\newblock \emph{IEEE Intelligent Systems}, 24\penalty0 (4):\penalty0 14 -- 20,
  2009.
\newblock ISSN 1541-1672.
\newblock \doi{10.1109/MIS.2009.69}.

\bibitem[O'~Meara(2012)]{OMeara2012}
Richard O'~Meara.
\newblock \emph{Robot Ethics:The Ethical and Social Implications of Robotics},
  chapter Contemporary Governance Architecture Regarding Robotics Technologies:
  An Assessment, page 400.
\newblock MIT Press, 2012.
\newblock ISBN 9780262298636.
\newblock URL
  \url{http://ieeexplore.ieee.org/xpl/articleDetails.jsp?arnumber=6733990}.

\bibitem[Russell(2015)]{Russell2015a}
Stuart Russell.
\newblock Ethics of artificial intelligence.
\newblock \emph{Nature}, 521\penalty0 (7553):\penalty0 415--416, MAY 28 2015.
\newblock ISSN 0028-0836.

\bibitem[Sharkey(2008)]{Sharkey2008}
Noel Sharkey.
\newblock The ethical frontiers of robotics.
\newblock \emph{Science}, 322\penalty0 (5909):\penalty0 1800--1801, 2008.

\bibitem[Stilgoe et~al.(2013)Stilgoe, Owen, and Macnaghten]{Stilgoe2013}
Jack Stilgoe, Richard Owen, and Phil Macnaghten.
\newblock Developing a framework for responsible innovation.
\newblock \emph{Research Policy}, 42\penalty0 (9):\penalty0 1568--1580, 2013.

\bibitem[Vanderelst and Winfield(2016)]{Vanderelst_submitted}
Dieter Vanderelst and Alan~FT Winfield.
\newblock An architecture for ethical robots.
\newblock Submitted, 2016.

\bibitem[Wallach and Allen(2008)]{Wallach2008}
Wendell Wallach and Colin Allen.
\newblock \emph{Moral machines: Teaching robots right from wrong}.
\newblock Oxford University Press, 2008.

\bibitem[Winfield(2011)]{Winfield2011}
Alan Winfield.
\newblock {Roboethics –for humans}.
\newblock \emph{New Scientist}, 210\penalty0 (2811):\penalty0 32--33, 2011.
\newblock ISSN 02624079.
\newblock \doi{10.1016/S0262-4079(11)61052-X}.
\newblock URL
  \url{http://linkinghub.elsevier.com/retrieve/pii/S026240791161052X}.

\bibitem[Winfield et~al.(2014)Winfield, Blum, and Liu]{Winfield2014}
Alan~FT Winfield, Christian Blum, and Wenguo Liu.
\newblock Towards an ethical robot: internal models, consequences and ethical
  action selection.
\newblock In \emph{Advances in Autonomous Robotics Systems}, pages 85--96.
  Springer, 2014.

\end{thebibliography}

\end{document}